\documentclass{article}
\usepackage{iclr2021_conference,times}

\usepackage{amsmath,amsfonts,bm}

\def\eqref#1{equation~\ref{#1}}

\def\1{\bm{1}}

\DeclareMathAlphabet{\mathsfit}{\encodingdefault}{\sfdefault}{m}{sl}
\SetMathAlphabet{\mathsfit}{bold}{\encodingdefault}{\sfdefault}{bx}{n}

\DeclareMathOperator{\sign}{sign}

\usepackage{hyperref}
\usepackage{url}

\usepackage{subcaption}
\usepackage{enumitem}
\usepackage{amsmath}
\usepackage{mathtools}
\usepackage{dcolumn}
\usepackage{siunitx}

\usepackage{hyperref}

\usepackage{amssymb}
\usepackage{tablefootnote}
\usepackage{multirow}
\usepackage{array}
\usepackage{booktabs}
\usepackage{graphicx}
\graphicspath{{./figures/}}

\usepackage{natbib}
\usepackage{caption}
\usepackage{lscape}
\usepackage{multicol}
\usepackage{stfloats}
\usepackage{xcolor}
\usepackage{algorithmic}
\usepackage{makecell}
\usepackage[T1]{fontenc}

\title{Sparse Coding Frontend for Robust Neural Networks}

\author{Can~Bakiskan, Metehan~Cekic, Ahmet~Dundar~Sezer, Upamanyu~Madhow\\
Department of Electrical and Computer Engineering\\
University of California Santa Barbara\\
Santa Barbara, CA, 93106, USA \\
\texttt{\{canbakiskan,metehancekic,adsezer,madhow\}@ece.ucsb.edu}
}

\iclrfinalcopy
\begin{document}

\maketitle

\begin{abstract}

Deep Neural Networks are known to be vulnerable to small, adversarially crafted, perturbations. The current most effective defense methods against these adversarial attacks are variants of adversarial training. 
In this paper, we introduce a radically different defense trained only on clean images: a sparse coding based front end which significantly attenuates adversarial attacks before they reach the classifier.
We evaluate our defense on CIFAR-10 dataset under a wide range of attack types (including $\ell^{\infty}$, $\ell^2$, and $\ell^1$ bounded attacks),
demonstrating its promise as a general-purpose approach for defense.

\end{abstract}


\section{Introduction}\label{sec:introduction}

Deep neural networks (DNNs) are known to be vulnerable to small, adversarially designed perturbations \citep{biggio2013evasion, szegedy2013intriguing}.
Since the discovery of such adversarial
attacks, researchers have tried various defense strategies such as adversarial training \citep{madry2017towards,zhang2019theoretically}, constraining Lipschitz constant \citep{cisse2017parseval}, randomized smoothing \citep{cohen2019certified}, and preprocessing methods \citep{guo2017countering,yang2019me}. Sadly, the majority of the proposed defenses have been subsequently defeated \citep{carlini2017adversarial,obfuscated-gradients}.  Current state-of-the-art defenses, as determined by several independent research efforts, are variants of adversarial training \citep{zhang2019theoretically,carmon2019unlabeled}. 

In this paper, we propose, and extensively evaluate on the CIFAR-10 image dataset, a novel preprocessing defense trained only on clean images.
We employ sparse coding using an overcomplete patch-level dictionary, and mitigate adversarial perturbations by (1) zeroing out all but the top $T$ coefficients of the basis projection, thus restricting the attack space, and
(2) quantizing the surviving coefficients to $\pm 1$, thus not allowing perturbations to ride on top of them.  The defense parameters are chosen to combat $\ell^{\infty}$ bounded
attacks, but we demonstrate its general-purpose nature by showing that it also provides robustness against $\ell^2$ and $\ell^1$ bounded attacks. The code for our defense can be found at: \url{https://github.com/canbakiskan/sparse_coding_frontend}


\section{Adversarial Attacks}

{\bf Whitebox attacks} assume that the adversary has access to the model and its weights. For typical $\ell^p$ bounded attack models, 
projected gradient descent (PGD) is currently known to be the strongest. The $\ell^p$ budget $\epsilon$ is divided into smaller steps $\delta$. In each step, gradient of the loss function $\mathcal{L}$ with respect to the input is normalized and scaled by $\delta$, and added to the perturbation calculated in the previous step. After each step, the computed attack is clipped to conform to the $\ell^p$ bound. Mathematically, PGD can be written as
\vspace*{-0.2cm}
\begin{equation}\label{eq:PGD}
\mathbf{e}_{i+1} = \text{clip}^p_{\epsilon}\big[\mathbf{e}_{i}+\delta \cdot \frac{\nabla_\mathbf{e} \mathcal{L}(\mathbf{f}(\mathbf{x}+\mathbf{e}_i),\mathbf{y})}{||(\nabla_\mathbf{e} \mathcal{L}(\mathbf{f}(\mathbf{x}+\mathbf{e}_i),\mathbf{y})||_p}\big]
\end{equation}
where $\mathbf{x}$ is the input, $\mathbf{y}$ is the target output, and $\mathbf{e}_i$ is the adversarial perturbation calculated in step $i$. The loss function $\mathcal{L}$ is usually taken as the cross-entropy loss between the network outputs $\mathbf{f}(\mathbf{x})$ and target output $\mathbf{y}$. Through \eqref{eq:PGD}, gradient ascent over the loss in the input space is achieved.

To optimize over a greater area in the input space, the attack can be initialized with random restarts within the $\epsilon$ $\ell^p$ ball of the clean input. After the computation of \eqref{eq:PGD} for each restart, a restart that successfully fools our model is chosen.

A modified version of PGD is called Carlini-Wagner (C\&W) attack. In this attack, the loss function $\mathcal{L}$ is taken as the difference between the logit that corresponds to the correct label and the maximum of logits among the incorrect labels. 

{\bf Blackbox attacks} do not assume full knowledge of the model and weights. Decision-based black box models assume access to the decision outputs of the model \citep{brendel2017decision},
while purely blackbox models assume only access to the training dataset, typically using a surrogate model (transfer blackbox) trained on the same dataset in order to generate attacks \citep{tramer2017space}.

\vspace*{-0.03cm}
\section{Frontend}\label{sec:frontend}

We consider RGB images of size $N \times N \times 3$, processed using  $n \times n \times 3$ patches with stride $S$, so that we process $M = m \times m$ patches, where $m =\lfloor{(N-n)/S}\rfloor +1$. 
We learn a patch-level overcomplete dictionary (number of basis tensors much larger than the patch dimension $3 n^2$) in unsupervised fashion, and use it to extract sparse local features. These outputs go through a CNN ``decoder'' whose goal is to restore the output
dimension to $N \times N \times 3$ rather than to reconstruct the image. This frontend is followed by a standard CNN classifier, trained jointly with the decoder using standard supervised learning. 

\vspace*{-0.03cm}
\subsection{Patch-level Overcomplete Dictionary}

We use a standard algorithm \citep{mairal2009online} (implemented in Python library \texttt{scikit-learn}) which is a variant of K-SVD \citep{elad2006image}. 
Given a set of clean training images $\mathcal{X}=\{\mathbf{X}^{(k)}\}_{k=1}^{K}$, an overcomplete dictionary $\mathbf{D}$ with $L$ atoms is learnt by solving the following optimization problem \citep{mairal2009online}:
\begin{equation}\label{eq:Dict}
\min _{\mathbf{D} \in \mathcal{C}, \{\boldsymbol{\alpha^{(k)}}\}_{k=1}^{K}} \sum_{k=1}^{K}\sum_{i,j}\Bigl( \frac{1}{2}\left\|\mathbf{R}_{ij}\mathbf{X}^{(k)}-\mathbf{D} \boldsymbol{\alpha}_{ij}^{(k)}\right\|_{2}^{2} +\lambda\left\|\boldsymbol{\alpha}_{ij}^{(k)}\right\|_{1}\Bigr)
\end{equation}
where $\mathcal{C} \triangleq \bigl\{\mathbf{D} = [\mathbf{d}_{1}, \ldots, \mathbf{d}_{L}] \in \mathbb{R}^{\bar{n} \times L} \mid \left\|\mathbf{d}_{l}\right\|_{2} = 1\,, \forall l \in \{1, \ldots, L \} \bigr\}$, $\lambda$ is a regularization parameter, $\boldsymbol{\alpha}^{(k)}$ is an $m \times m \times L$ tensor containing the coefficients of the sparse decomposition, and $\mathbf{R}_{ij} \in \mathbb{R}^{\bar{n} \times \bar{N}}$ with $\bar{n} \triangleq 3n^2$ and $\bar{N} \triangleq 3N^2$ extracts the $(ij)$-th patch from image $\mathbf{X}^{(k)}$. The optimization problem in \eqref{eq:Dict} is then solved for the two variables $\mathbf{D}$ and $\{\boldsymbol{\alpha}^{(k)}\}_{k=1}^{K}$ in an alternating fashion \citep{mairal2009online,elad2006image}.

The learnt dictionary atoms form the first layer of our frontend, so that the first layer outputs are the projections of the patches onto the atoms.
Specifically, for a given image $\mathbf{X}$, patch $\mathbf{x}_{ij} \in \mathbb{R}^{\bar{n}}$ is extracted based on the $(ij)$-th block of $\mathbf{X}$; that is, $\mathbf{x}_{ij} = \mathbf{R}_{ij}\mathbf{X}$, and then projected onto dictionary $\mathbf{D}$ in order to obtain projection vector $\mathbf{\tilde{x}}_{ij}$, where $\mathbf{\tilde{x}}_{ij} = \mathbf{D}^T\mathbf{x}_{ij}$. 

\begin{figure}[b]
\centering
\begin{subfigure}{.545\textwidth}
  \centering
  \includegraphics[width=.97\linewidth]{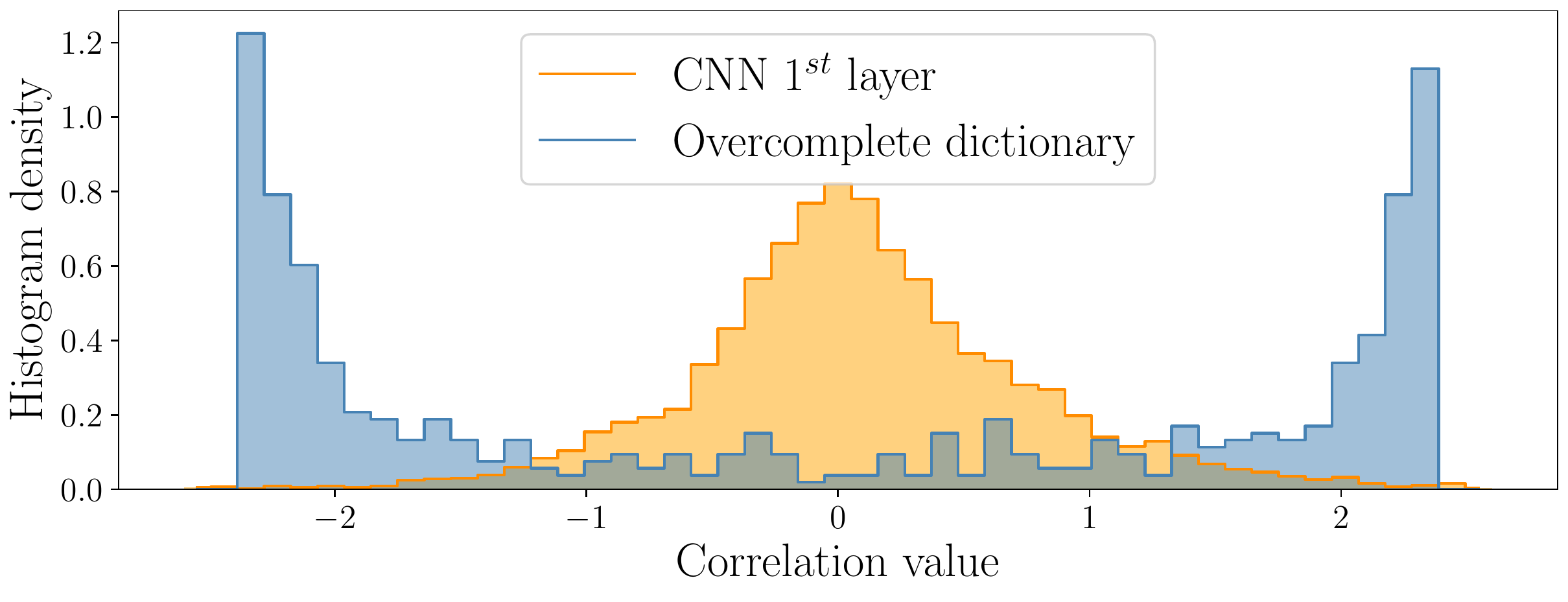}
  \caption{}
  \label{fig:correlations}
\end{subfigure}%
\begin{subfigure}{.455\textwidth}
  \centering
  \includegraphics[width=.97\linewidth]{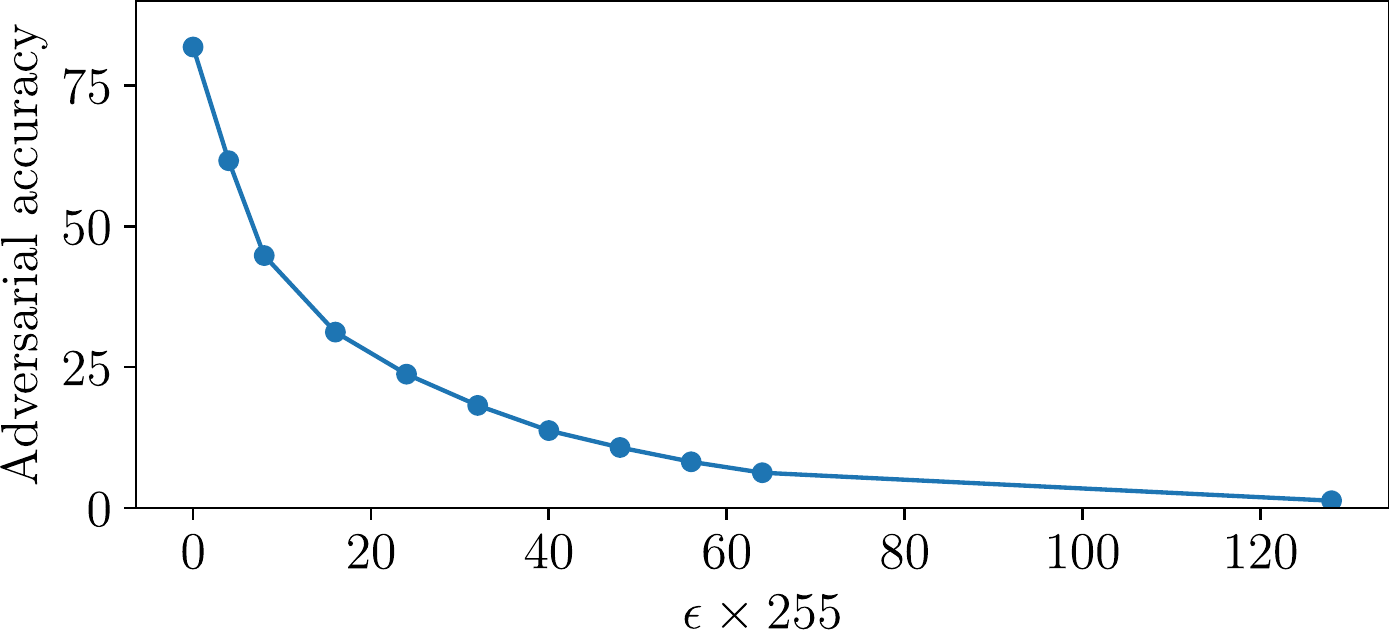}
  \caption{}
  \label{fig:acc_vs_eps}
\end{subfigure}
\caption{(a) Histogram of correlations for a typical patch with atoms of an overcomplete dictionary vs. that of activations through layer 1 filters of a standard classifier CNN. (b) Accuracy vs $\epsilon$ plot for our defense ($T=15$) under $\ell^\infty$ PGD attack with various $\epsilon$ values.}
\label{fig:ensemble_benefit}
\end{figure}

Since the dictionary is highly overcomplete, the majority of coefficients are relatively small, and a sparse representation of the patch can be obtained from a subset of the coefficients taking large values.
Fig.~\ref{fig:correlations} contrasts the typical distribution of coefficients from our dictionary to that for the first layer for a typical CNN (the latter is not amenable to sparse coding).

\vspace*{-0.03cm}

\subsection{Coefficient Selection and Activation}\label{sec:topT_activation}

For each patch, we keep only the $T$ elements of the projection vector with largest absolute values, zeroing out the remaining elements, limiting the dimension of the subspace available to the adversary. The surviving coefficients are denoted by $\mathbf{\hat{x}}_{ij}$. Next, in order to ensure that the adversarial perturbations cannot ride on top of the signal in the surviving coefficients $\mathbf{\hat{x}}_{ij}$, we apply an activation function, zeroing out coefficients smaller than a threshold, and keeping only the sign of coefficients passing the threshold. 
The specific design we consider targets $\ell^{\infty}$ bounded perturbations of size $\epsilon$ (but as we shall see, provides robustness against $\ell^2$ and $\ell^1$ bounded perturbations as well):
we set the threshold for atom $\mathbf{d}_{l}$  to be a multiple of $\epsilon \left\|\mathbf{d}_{l}\right\|_{1}$, the maximum possible contribution of 
of such a perturbation to the output. 
\begin{equation}\label{eq:act_fun}
\mathbf{\bar{x}}_{ij}(l)=\biggl\{\begin{array}{ll}
\sign{(\mathbf{\hat{x}}_{ij}(l))}\left\|\mathbf{d}_{l}\right\|_{1}, & \text { if } \frac{|\mathbf{\hat{x}}_{ij}(l)|}{\epsilon \left\|\mathbf{d}_{l}\right\|_{1}} \geq \beta\\
0, & \text{otherwise}
\end{array},
\end{equation}
for all $l \in\{1, \ldots, L\}$, where $\beta > 1$ is a hyperparameter. The scaling of the surviving $\pm 1$ outputs by $\left\|\mathbf{d}_{l}\right\|_{1}$ 
allows basis coefficients surviving a larger $\ell^1$ norm based threshold to contribute more towards the decoder input, but could be omitted, since the decoder can learn the appropriate weights.

\subsection{Decoder}

The discretized sparse codes $\mathbf{\bar{x}}_{ij}$ are fed into a CNN decoder with three transposed convolution layers followed by ReLU activations after each layer, which restores the original image size.

\section{Evaluation}\label{sec:evaluation}

State-of-the-art PGD attacks are based on gradient computation, and an important criticism of prior proposals of preprocessing-based defense is that they were masking/obfuscating/shattering 
gradients, without adapting the attacks to compensate for this \citep{obfuscated-gradients}. Indeed, many such defenses do fail when attacks are suitably adapted, leading a group of prominent
researchers to propose guidelines for evaluating defenses \citep{onevaluating2019}. Following such guidelines, we devote substantial effort to devising attacks adapted to our defense as follows:
\begin{itemize}
	\item Consider several different smooth backward pass replacements for the parts of our defense that are non-differentiable:
		\begin{itemize}
			\item for the activation function, we test two backward passes: identity and a smooth approximation. 
			\item for top $T$ taking operation, we test identity backward pass as well as propagating gradients through top $U$ coefficients where $U>T$.
		\end{itemize}
	\item Check that there exist some $\epsilon$ values that reduce the adversarial accuracy to 0. (see Fig.~\ref{fig:acc_vs_eps})
	\item Test our defense against attacks with large number of iterations or restarts.
	\item Test against different threat models such as: $\ell^\infty$, $\ell^2$, and $\ell^1$ bounded attacks; decision and query based blackbox attacks that do not rely on gradients; transfer blackbox attacks with many different surrogate models
	\item To ensure that the gradients are computed appropriately, we test computing smoothed gradients where gradients are averaged over many different points in the neighborhood of the original point, in each step of the attack. However, we observe that this does not significantly increase the attack strength.
\end{itemize}


\section{Experiments and Results}

\renewcommand{\arraystretch}{1.3}
\setlength{\tabcolsep}{3pt}
\begin{table}[!b]\centering
\begin{small}
\begin{tabular}{r@{\hspace{15pt}}ccccccc}
\toprule

& Clean & \makecell{Whitebox \\ ($\ell^\infty$, $\epsilon=\frac{8}{255}$)}& \makecell{Whitebox (C\&W) \\ ($\ell^\infty$, $\epsilon=\frac{8}{255}$)}  & \makecell{Whitebox \\ ($\ell^2$, $\epsilon=0.6$)}   &  \makecell{Whitebox \\ ($\ell^1$, $\epsilon=30$)}  & \makecell{Decision \\ Boundary BB}    \\

\midrule
Natural & \bf{92.66} & \phantom{0}0.00 & \phantom{0}0.00 & \phantom{0}0.00 & \phantom{0}0.00 & $\overline{||\mathbf{e}||_2}=0.11$\\
Adv. Training & 79.41 & 43.27 & 42.03 & 52.09 & 19.98 & $\overline{||\mathbf{e}||_2}=0.88$\\
TRADES & 75.17 & \bf{45.79} & \bf{42.87} & 51.35 & 21.15 & $\overline{||\mathbf{e}||_2}=0.91$\\
$T=15$ (Ours) & 85.45 & 37.33 & 37.35 & \bf{60.47} & \bf{47.13} & $\overline{||\mathbf{e}||_2}=\bf{3.04}$\\

\bottomrule
\end{tabular}
\end{small}
\caption{Performance (accuracies and norms) of different defense methods (CIFAR-10) }
\label{table:different_defenses}
\end{table} 

We test our defense against a wide range of attacks and compare the results with benchmark defenses from the literature using the CIFAR-10 dataset (pixels scaled between 0 and 1). The benchmarks are PGD adversarially trained \citep{madry2017towards} and TRADES \citep{zhang2019theoretically} defenses for the same classifier architecture under $\ell^{\infty}$ bounded attacks. See Appendix~\ref{sec:hyper} for choice of hyperparameters and training settings for both our defense and benchmarks. Note that the classifier CNN used in our paper is "simpler" ResNet-32 rather than the wide ResNet-32, both of which are utilized in \cite{madry2017towards} and other studies in the literature. The choice of the smaller ResNet-32 network makes evaluation of stronger attacks computationally more feasible.

Table~\ref{table:different_defenses} compares our defense with the benchmarks from the literature (all columns except the last report classification accuracies). In all whitebox attacks, we use attack settings: number of steps $N_s=40$, number of random restarts $N_r=100$ and step size $\delta=\epsilon/8$. For the attacks against our defense, we also use $U=2T$ and backward pass smoothing mentioned in Section~\ref{sec:evaluation} to make it stronger. See Appendix~\ref{sec:attack_details} for discussion of these choices and accuracies for other attack variations. Column~1 reports on clean accuracy. In columns 2 and 3, we report results for the $\ell^\infty$ bounded whitebox attack with cross entropy loss and Carlini-Wagner loss, respectively.  
In columns 4 and 5, we report results for $\ell^2$ and $\ell^1$ bounded attacks, respectively. For the last (sixth) column, we use a decision-based blackbox attack \citep{brendel2017decision} and report the average $\ell^2$ norm of perturbations that brings the classification accuracy to $\sim0.01\%$. This attack does not rely on gradient computation. For a given datapoint, the starting point for the perturbation search in this attack is chosen as the closest (in the $\ell^2$ sense) datapoint with a different label. Other parameters for this attack are the default parameters in \texttt{foolbox} \citep{rauber2017foolboxnative} Python package.

We highlight the following observations from Table~\ref{table:different_defenses}: (a) the clean accuracy (column 1) is better than that of the benchmarks, (b) the adversarial accuracy (columns 2 and 3) for $\ell^{\infty}$ bounded perturbations falls short of the benchmarks, (c) the adversarial accuracy for $\ell^2$ (column 4) and $\ell^1$ (column 5) bounded perturbations is substantially better than that of the benchmarks, (d) the average $\ell_2$ norm (column 6) required to create a classification error with high probability with a non-gradient attack is substantially higher for our defense. 

The robustness of our defense to $\ell^{p}$ attacks with different $p$ is because the impact of different attacks is similar: they flip $\pm1$ to $0$ and $0$ to $\pm1$ in the sparse code.
In contrast, standard adversarial training approaches are optimized for $\ell^p$ bounded attacks for a specific $p$, and do not perform as well under other types of attacks.


\section{Conclusion}

The promising results obtained with our sparse coding front end open up an exciting new direction for building bottom-up defenses to make neural networks more robust for a wider range of attacks: we can tune our defense to approach the performance of state-of-the-art adversarial training for $\ell^{\infty}$ bounded attacks, while also providing robustness against $\ell^2$ and $\ell^1$ bounded attacks.  The ideas that our defense is based on could be applied to not only the first layer, but also to intermediate layers within neural networks. Such an approach could distribute the workload of combatting attacks across layers, and potentially make the networks more robust to a variety of attacks from within, without the need for adversarial training adapted to a specific class of attacks.


\subsection*{Acknowledgment}

This work was supported in part by the Army Research Office under grant W911NF-19-1-0053, and by the National Science Foundation under grants CIF-1909320 and CNS-1518812.

\newpage
\bibliography{references}
\bibliographystyle{iclr2021_conference}

\appendix

\section{Appendix}\label{sec:appendix}

\subsection{Block Diagram of the Defense}

\begin{figure}[!ht]
	\centering
 	\includegraphics[width=.6\linewidth]{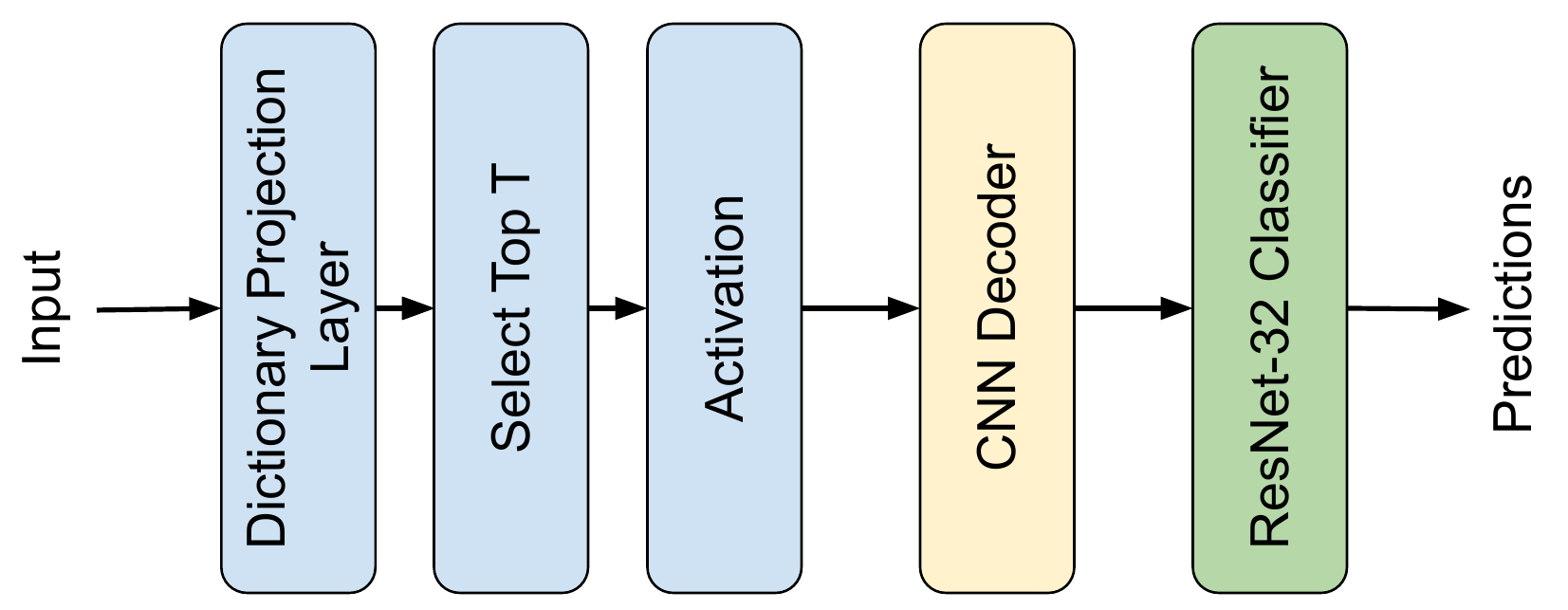}
 	\caption{A block diagram of our proposed defense}
 	\label{fig:pipeline}
\end{figure}

Input images first go through the convolutional layer with the dictionary atoms being its filters. Then, for each patch representation, Top $T$ coefficients in absolute value are kept while other coefficients are zeroed out. This is followed by the activation function described in Section \ref{sec:topT_activation}. These quantized and highly sparse representations are then fed into a convolutional neural net based decoder which restores the image size. The outputs of the frontend are the inputs of the ResNet-32 based classifier. Training of the decoder and the classifier are done simultaneously after the dictionary is learnt and frozen.

\subsection{Determining Attack Parameters} \label{sec:attack_details}

In order to determine the strongest attack settings against our defense, we test it with different attack parameters. In Table~\ref{table:ours_comprehensive}, we report five whitebox attack settings on our defense for different $T$ values. For these attacks, every differentiable operation is differentiated. For the non-differentiable activation function, we take a smooth backward pass approximation. We also experiment with replacing it with identity in the backward pass but this results in weaker attacks (see \ref{sec:bpda}). For the baseline attack settings, we choose number of restarts $N_r=10$, number of steps $N_s=40$, and step size $\delta=1/255$. We then change one or two settings at a time and report those in each column. In the third column, we report accuracies when gradients are propagated through the top $U$ coefficients (rather than top $T$). For the fourth column, we increase $N_s$ to $1000$, reduce $\delta$ to $0.5$ and $N_r$ to $1$ in order to keep computational complexity reasonable. For the fifth column, we increase $N_r$ to $100$ and keep $N_s=40$. Finally, for sixth column, we propagate the gradients through the top $U$ coefficients as well as increasing the number of restarts $N_r$ to 100. For all values of $T$, last two columns' settings result in the strongest attacks. Increasing the number of restarts $N_r$ to 100 especially, had the biggest impact on the accuracies. 

\renewcommand{\arraystretch}{1.2}
\setlength{\tabcolsep}{6pt}
\begin{table}[!ht]\centering
\begin{small}
\begin{tabular}{r@{\hspace{15pt}}cccccc}
\toprule

& Clean & \makecell{Whitebox \\ (Baseline)}& \makecell{Whitebox \\ ($U=2T$)} & \makecell{Whitebox \\ ($N_s=1000$)}   & \makecell{Whitebox \\ ($N_r=100$)} & \makecell{Whitebox \\ ($U=2T$, $N_r=100$)}\\

\midrule
$T=1$ & 81.81 & 44.79 & 48.26 & 59.91 & 34.89 & 37.83 \\
$T=2$ & 83.03 & 47.89 & 47.81 & 61.35 & 37.48 & 37.92 \\
$T=5$ & 83.24 & 47.11 & 42.21 & 60.37 & 36.66 & 32.83 \\
$T=10$ & 84.71 & 48.48 & 44.83 & 60.06 & 39.69 & 36.16 \\
$T=15$ & 85.45 & 46.89 & 44.55 & 57.13 & 38.43 & 37.33 \\
$T=20$ & 85.46 & 40.96 & 45.27 & 51.39 & 32.63 & 37.25 \\

\bottomrule
\end{tabular}
\end{small}
\caption{Accuracies for our defense method under different settings (CIFAR-10, $\ell^\infty$ $\epsilon=8/255$) }
\label{table:ours_comprehensive}
\end{table}

By looking at the results in Table~\ref{table:ours_comprehensive}, we determine to report on $T=15$ in Table~\ref{table:different_defenses}, since it represents a good tradeoff between clean and attacked accuracy. After the decision to use $T=15$ for benchmark comparison in Table~\ref{table:different_defenses}, we use the strongest attack for this defense setting, which is taking $N_r=100$ and propagating gradients through the top $U=2T=30$ coefficients.

\subsection{Choice of Hyperparameters and Training Settings}\label{sec:hyper}

\noindent {\it Our defense:} We evaluate our defense on the CIFAR-10 dataset ($N=32$), for which there are well-established benchmarks in adversarial ML. In our defense, we use $4 \times 4$ patches ($n=4$) and an overcomplete dictionary with $L=500$ atoms. The stride $S=2$, so the encoder output is a $15 \times 15 \times 500$ tensor ($m=15$, $L=500$). The regularization parameter in \eqref{eq:Dict} is set to $\lambda = 1$, in the upper range of values resulting in convergence. The number of iterations in dictionary learning is chosen as $1000$ to ensure convergence. The number of dictionary atoms $L$  is chosen to be $10$ times the ambient dimension of patches. 

We test our defense for $T=1,2,5,10,15,$ and $20$ with hyperparameter $\beta = 3$ for the threshold in \eqref{eq:act_fun}. We train the CNN-based decoder in supervised fashion in tandem with the classifier, using the standard cross-entropy loss. We use a cyclic learning rate scheduler \citep{smith2017cyclical} with a maximum learning rate of $\eta_{max}=0.05$ for $T=1,2$ and $\eta_{max}=0.02$ for $T=5,10,15,20$. In order to provide a consistent evaluation, we employ the ResNet-32 classifier used in \cite{madry2017towards} and train it for $70$ epochs.

\noindent {\it Benchmarks:} For a fair comparison, we use the same ResNet-32 classifier architecture for the benchmarks. We train the PGD adversarially trained model from \cite{madry2017towards} with the same cyclic learning rate with $\eta_{max}=0.05$ for $100$ epochs. We train the model for TRADES defense with learning rate $\eta=0.01$ for the first 50 epochs and then with $\eta=10^{-3}$ for the next 50 epochs. For both PGD adversarially trained model and TRADES, training hyperparameters are $\epsilon=8/255$, $\delta=1/255$, $N_S=10$, $N_R=1$. Additionally for TRADES $\lambda_{\text{TRADES}}=1/6$. We also report on naturally trained network (i.e., no defense). This network is also trained for $70$ epochs with the same cyclic learning rate with $\eta_{max}=0.05$.

\subsection{Backward Pass Approximation to Activation}\label{sec:bpda}

\begin{figure}[!ht]
	\centering
 	\includegraphics[width=.75\linewidth]{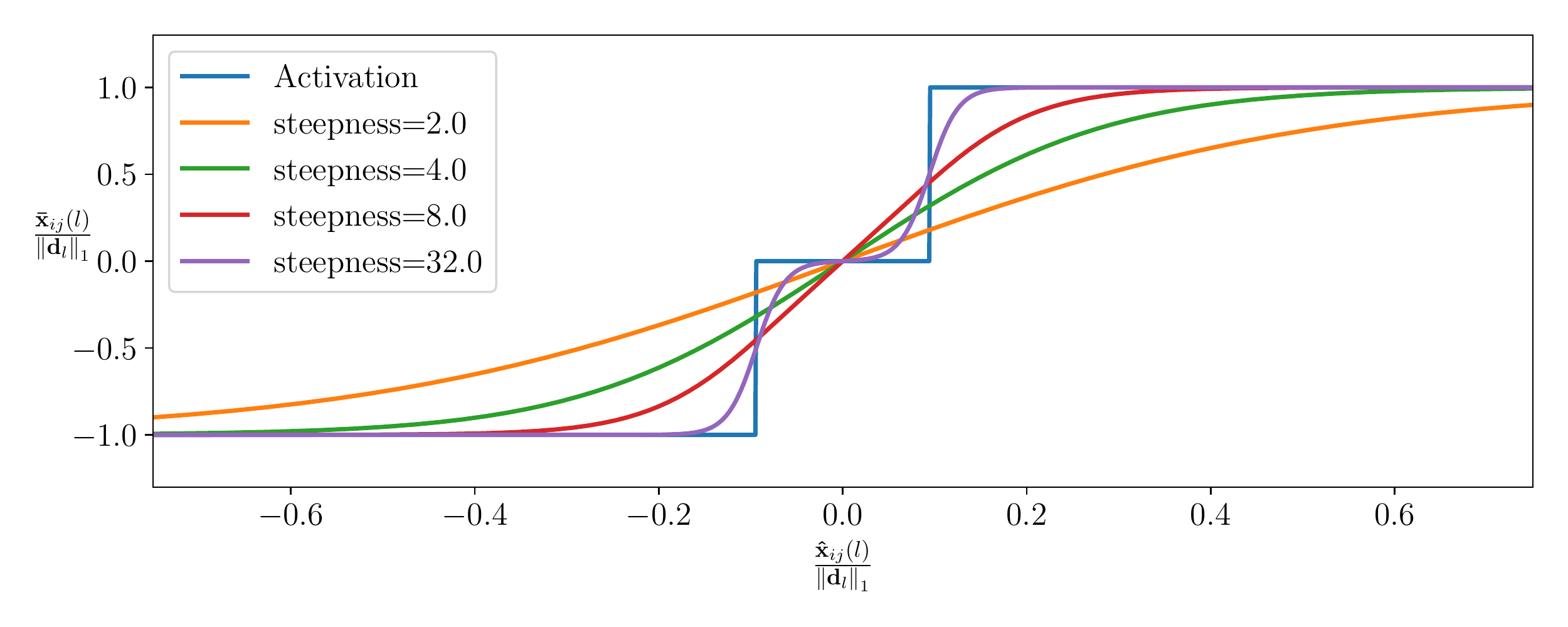}
 	\caption{Activation function and its backward pass smooth approximation with varying degrees of smoothness}
 	\label{fig:bpda}
\end{figure}

We try replacing the activation function with two different functions in the backward pass: identity and a smooth approximation (\ref{fig:bpda}). We observe that the approximation with steepness$=4.0$ results in the strongest attacks and use it in the backward pass of all reported attacks.

\subsection{Gradient Propagation Through Top U Coefficients}

We test different settings for the hyperparameter $U$ such as $U=2T$, $U=5T$, $U=10T$ and find that, in general, $U=2T$ results in the strongest attacks. 

\subsection{Gradient Smoothing}

We test smoothing the gradients in each step by averaging them over 40 different points in the $\delta$ $\ell^\infty$ neighborhood in each step of the attack. For computational feasibility, we reduce $N_s$ to $20$ and $N_r$ to $1$. In general, for the models we tried this on, this smoothing operation did not result in significantly stronger attacks. It decreased accuracies less than $0.5\%$ while increasing the computational cost significantly. For this reason, we choose not to use this operation in the attacks reported.

\subsection{Results for HopSkipJumpAttack}

HopSkipJumpAttack \citep{chen2020hopskipjumpattack} is a recently proposed decision-based attack. The results for this attack are reported in Table~\ref{table:hopskip}. Similar to the last column of Table~\ref{table:different_defenses} the reported results are for the average $\ell^2$ norm of the perturbations. We observe that, like the decision boundary attack, our defense requires the highest average $\ell^2$ norm perturbations.

\renewcommand{\arraystretch}{1.3}
\setlength{\tabcolsep}{6pt}
\begin{table}[!htb]\centering
\begin{small}
\begin{tabular}{r@{\hspace{15pt}}c}
\toprule
\multicolumn{2}{c}{HopSkipJumpAttack}     \\
\midrule
& Average $\ell^2$ norm of successful attack images\\
\midrule
Natural & $\overline{||\mathbf{e}||_2}=0.31$\\
Adv. Training & $\overline{||\mathbf{e}||_2}=2.08$\\
TRADES & $\overline{||\mathbf{e}||_2}=2.00$\\
$T=15$ (Ours) & $\overline{||\mathbf{e}||_2}=\bf{3.43}$\\

\bottomrule
\end{tabular}
\end{small}
\caption{Results for HopSkipJumpAttack attack (CIFAR-10) }
\label{table:hopskip}
\end{table} 

\subsection{Results for Zeroth Order Optimization (ZOO) Based Attack}

ZOO attack \citep{zoo2017} is a query-based blackbox attack where the gradients of the model with respect to each input pixel are approximated using numerical differentiation and then the attack is computed through stochastic coordinate descent. For computational complexity reasons we evaluate this attack on 100 randomly selected images. These do not include images that are wrongly classified by the corresponding model. The results for this attack are reported in Table~\ref{table:zoo}. Similar to the last column of Table~\ref{table:different_defenses} the reported results are for the average $\ell^2$ norm of the perturbations. We observe that the ZOO attack is unable to find adversarial examples for our defense. This is due to the attack computing gradients by changing one pixel at a time by $\pm \Delta x$. Since our frontend is very insensitive to single pixel changes, the gradients cannot be calculated using such numerical differentiation techniques. We acknowledge that while this observation alone doesn't show that our defense is secure, it does mean that this particular type of attack is not applicable to our defense.

\renewcommand{\arraystretch}{1.3}
\setlength{\tabcolsep}{6pt}
\begin{table}[!hb]\centering
\begin{small}
\begin{tabular}{r@{\hspace{15pt}}cc}
\toprule
\multicolumn{3}{c}{ZOO Attack}     \\
\midrule
& Attack success rate (\%) & Average $\ell^2$ norm of successful attack images\\
\midrule
Natural & 100 & $\overline{||\mathbf{e}||_2}= 0.13$\\
Adv. Training & 100 & $\overline{||\mathbf{e}||_2}=0.89$ \\
TRADES & 100 & $\overline{||\mathbf{e}||_2}=0.93$ \\
$T=15$ (Ours) & 7 & $\overline{||\mathbf{e}||_2}=0.04$\\

\bottomrule
\end{tabular}
\end{small}
\caption{Results for ZOO attack (CIFAR-10) }
\label{table:zoo}
\end{table}

\subsection{Effect of Attack Step Size}

Nominally, we report results for when the stepsize $\delta=\epsilon/8$. For the $\ell^\infty$ attack, when we use $N_s=1000$, we decrease $\delta$ to $0.5/255$. We also test for $\delta=2/255$ but this results in higher accuracies therefore is omitted in the tables. For the $\ell^1$ and $\ell^2$ bounded attacks, we try $\delta=\epsilon/20$ but this too results in weaker attacks.

\subsection{Validation of Attack Code}

To generate all attacks except for the decision-based blackbox attacks, we use our own attack library and validate our results with the \texttt{foolbox} \citep{rauber2017foolboxnative} and \texttt{torchattacks} \citep{kim2020torchattacks} Python packages. To generate the decision based blackbox attacks, we use the \texttt{foolbox} package. 

\end{document}